\documentclass[letterpaper, 10 pt, conference]{ieeeconf}  

\IEEEoverridecommandlockouts                              
\usepackage[utf8]{inputenc}
\usepackage{amsmath,amssymb,stmaryrd,mathtools}
\usepackage{amsfonts}
\usepackage[T1]{fontenc}
\usepackage[noend]{algpseudocode}
\usepackage{acronym}
\usepackage{xcolor}
\usepackage{verbatim}
\usepackage{booktabs}
\usepackage{siunitx}
\usepackage{graphics}
\usepackage{graphicx, caption}
\usepackage{flushend}
\usepackage{suffix}
\usepackage{xstring}
\usepackage{xparse}
\usepackage{expl3}
\usepackage{mathrsfs}
\usepackage{tabularx}
\usepackage{makecell}
\usepackage{array}
\usepackage{hyperref}

\usepackage{cleveref}
\usepackage[ruled,linesnumbered]{algorithm2e}
\usepackage{multirow}
\usepackage{diagbox}
\usepackage{rotating}

\usepackage{subcaption}
\usepackage{wrapfig}

\usepackage{tikz}
\usetikzlibrary{arrows,backgrounds,calc}
\usepackage{relsize}
\usepackage{float}
\usepackage{kantlipsum} 
\usepackage{lipsum}
\usepackage{stfloats}
\usepackage{siunitx}

\usepackage{xcolor}

\usepackage[noadjust]{cite}
\usepackage{url}

\usepackage{todonotes}
\usepackage{soul}
\definecolor{smoothgreen}{rgb}{0.7,1,0.7}
\sethlcolor{smoothgreen}

\RequirePackage{luatex85}
\usepackage{pgfplots}
\pgfplotsset{compat=newest}
\pgfplotsset{every axis legend/.append style={%
		cells={anchor=west}}
}
\usepgfplotslibrary{polar}
\usetikzlibrary{arrows}
\tikzset{>=stealth'}

\definecolor{C1}{rgb}{0.0, 0.447, 0.741}
\definecolor{C1_light}{rgb}{0.0, 0.6032388663967612, 1.0}
\definecolor{C2}{rgb}{0.85, 0.325, 0.098}
\definecolor{C3}{rgb}{0.929, 0.694, 0.125}
\definecolor{C4}{rgb}{0.494, 0.184, 0.556}
\definecolor{C5}{rgb}{0.466, 0.674, 0.188}
\definecolor{C6}{rgb}{0.301, 0.745, 0.933}
\definecolor{C7}{rgb}{0.635, 0.078, 0.184}

\usepgfplotslibrary{groupplots}

\usetikzlibrary{shapes.geometric, arrows}

\tikzstyle{startstop} = [rectangle, rounded corners, minimum width=2cm, minimum height=1cm,text centered, draw=black, fill=none]
\tikzstyle{arrow} = [thick,->,>=stealth]

\title{\LARGE \bf 
Multi-Modal Manipulation via Multi-Modal Policy Consensus   
}

\author{%
Haonan Chen$^{1,4}$,
Jiaming Xu$^{1,*}$,
Hongyu Chen$^{1,*}$,
Kaiwen Hong$^1$,
Binghao Huang$^2$,
Chaoqi Liu$^1$, \\
Jiayuan Mao$^3$,
Yunzhu Li$^2$,
Yilun Du$^{4,+}$,
Katherine Driggs-Campbell$^{1,+}$ \\
\\
$^1$ University of Illinois Urbana-Champaign \quad
$^2$ Columbia University \\
$^3$ Massachusetts Institute of Technology \quad
$^4$ Harvard University \\
$^*$ Equal contribution \quad
$^+$ Equal advising\\ 
\\
\normalsize \href{https://policyconsensus.github.io}{https://policyconsensus.github.io}
}

\begin{document}
\twocolumn[{%
	\renewcommand\twocolumn[1][]{#1}%
    \maketitle
    \vspace{-6mm}
    \centering
    \includegraphics[width=0.99\linewidth]{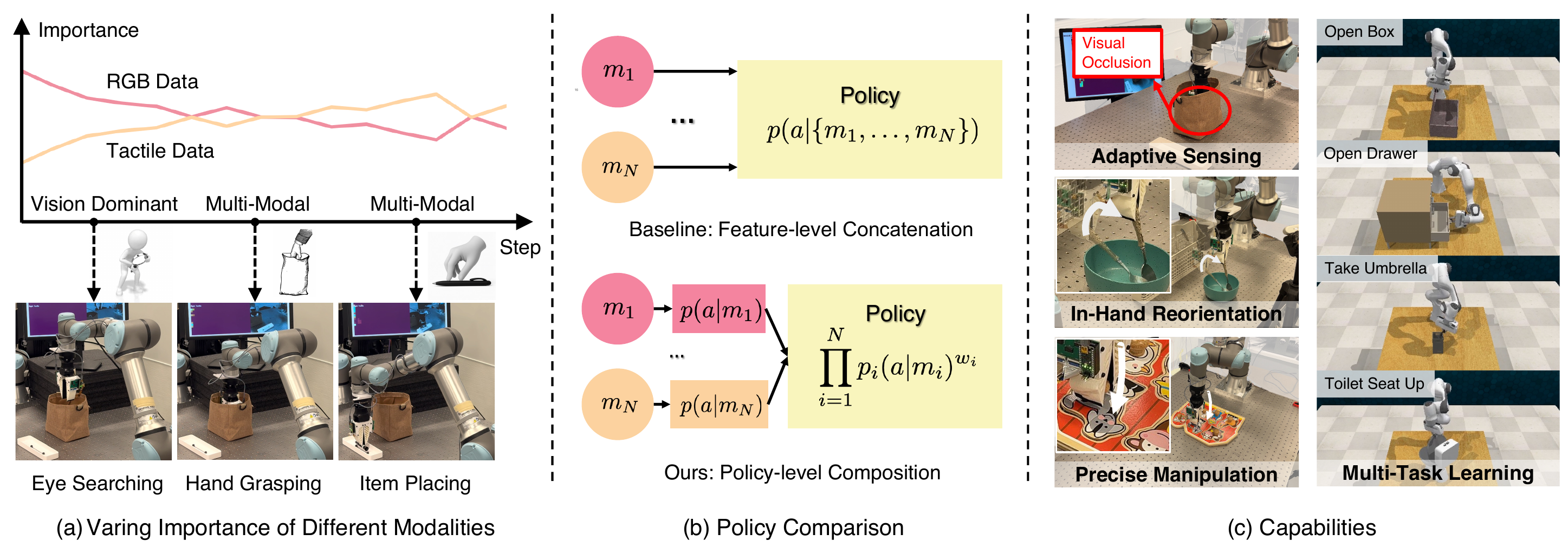}
    \captionof{figure}{
    \textbf{Representation-Composable Policy.} 
    (a) Perturbation-based importance analysis in the occluded marker picking task shows that vision dominates early, while tactile signals become important once occluded, demonstrating that our framework dynamically utilizes different modalities across task phases. 
    (b) Classical feature concatenation vs.~our policy-level composition, where $m_i$ denotes a modality (e.g., RGB, point cloud, tactile, or learned visual feature). Our compositional design allows individual modality policies to be added or removed without retraining the entire network. 
        (c) Our method unlocks key capabilities. 
    These include \textit{Adaptive Sensing}, retrieving an occluded marker using tactile feedback during occlusion; 
    \textit{In-Hand Reorientation}, reorienting a spoon within the gripper; 
    \textit{Precise Manipulation}, inserting a puzzle piece with fine-grained control; 
    and \textit{Multi-Task Learning}, consistently outperforming prior work across diverse tasks in {RLBench}.}
        
    \label{fig:teaser}
}]

  

\begin{abstract}
Effectively integrating diverse sensory modalities is crucial for robotic manipulation. However, the typical approach of feature concatenation is often suboptimal: dominant modalities such as vision can overwhelm sparse but critical signals like touch in contact-rich tasks, and monolithic architectures cannot flexibly incorporate new or missing modalities without retraining. 
Our method factorizes the policy into a set of diffusion models, each specialized for a single representation (e.g., vision or touch), and employs a router network that learns consensus weights to adaptively combine their contributions, enabling incremental of new representations.
We evaluate our approach on simulated manipulation tasks in {RLBench}, as well as real-world tasks such as occluded object picking, in-hand spoon reorientation, and puzzle insertion, where it significantly outperforms feature-concatenation baselines on scenarios requiring multimodal reasoning. Our policy further demonstrates robustness to physical perturbations and sensor corruption. We further conduct perturbation-based importance analysis, which reveals adaptive shifts between modalities. 
Project website: https://adaptivescene.github.io
 \looseness=-1



\end{abstract}

\section{Introduction}

Modern robots utilize a diverse array of modalities including RGB images, point clouds, tactile signals, and learned visual features, yet effectively integrating these data streams remains a challenge in robotics~\cite{chang2020sound, lee2019multimodal}. These modalities can be both complementary (vision vs. touch) and overlapping (RGB-D vs. point cloud), requiring structured synergy for optimal performance. For decades, a common baseline has been the concatenation of feature-level modalities into a single high-dimensional vector, an approach that persists even in recent policies~\cite{lee2019multimodal, johanna2022multimodal}. Despite its popularity, this approach lacks a principled mechanism for balancing contributions across modalities and cannot easily adapt when modalities are added or missing, often resulting in suboptimal performance.


The weakness becomes apparent when different sensors are crucial at different task phases. Consider a robot trained to retrieve an object from an opaque container. For 90\% of the trajectory, vision guides the robot to approach and position its gripper. Once the gripper enters the container, however, vision becomes useless while tactile signals become critical for success. Despite tactile information being essential during this crucial 10\% of the task, feature concatenation struggles with this sparsity: the learning algorithm downweights the rarely-active tactile stream as noise, focusing instead on the statistically dominant visual features. Moreover, these monolithic approaches cannot adapt to new sensors or missing modalities without complete retraining. These limitations point to the need for a structured alternative that treats each modality as a distinct contributor rather than forcing premature fusion.

To address these challenges, we draw inspiration from compositional generative models~\cite{du2020compositional, liu2022compositional, du2024reducereuserecyclecompositional, du2024compositional} and factorize robot policies into \textit{modality-specific experts}, with a router network learning consensus among them. Each expert specializes in a single modality (e.g., vision for geometric reasoning or tactile for contact dynamics) and captures its behavioral constraints~\cite{yang2023compositional}. A router network learns consensus weights during training to combine these experts into a unified policy, ensuring that even rarely-active but crucial modalities (e.g., tactile) retain their influence when most needed. As illustrated in Figure~\ref{fig:teaser}b, our approach contrasts with classical feature-level concatenation by composing experts at the policy level, which not only mitigates sparsity issues but also allows new modalities to be added or removed without retraining the entire network. This compositional design directly resolves the limitations of prior work by preserving the contribution of each modality, adapting flexibly to sensor availability, and providing robustness to failures, while also enabling diverse manipulation skills such as those shown in Figure~\ref{fig:teaser}c. \looseness=-1

Building on this compositional structure, our framework enables context-aware shifts in modality reliance: tactile experts dominate during contact-rich phases, while vision governs geometric reasoning in free space. The unified policy, formed by weighted sums of score functions from the factorized diffusion experts, yields robustness under sensor corruption and partial observability, ensuring reliable manipulation across diverse conditions.


Our contributions are as follows:
\begin{enumerate}
    \item We introduce a framework that composes {modality-specific experts} through learned consensus weights, offering a principled and extensible alternative to monolithic feature concatenation. 
    This design naturally supports incremental learning, allowing new experts to be integrated without retraining existing policies.
    \item We demonstrate strong performance on the multi-task RLBench~\cite{james2019rlbenchbenchmark} simulation benchmark compared to various baselines and validate our approach on complex real-world tasks, including occluded marker picking, spoon reorientation in hand, and puzzle insertion. We also show robustness to physical perturbations, runtime disturbances, and sensor corruption.
    \item We provide comprehensive analyses, including perturbation-based importance studies, that quantitatively demonstrate how our policy learns to shift reliance between modalities in response to changing task context.
\end{enumerate}
\section{Related Works}

\noindent\textbf{Multimodal Fusion in Robotics.} 
The integration of vision and touch is critical for robust robotic manipulation, allowing robots to handle occlusions and make precise contact~\cite{Johansson2004RolesOG, Bicchi2000RoboticGA, Huang2023TowardsSM}. Vision provides global scene understanding while tactile sensing offers high-fidelity local information critical for tasks like grasping and insertion~\cite{suresh2023neural, dave2024multimodal}. Existing fusion approaches range from simple feature concatenation~\cite{lee2019multimodal, li2023see} to sophisticated architectures including Visuo-Tactile Transformers that dynamically weigh modality features~\cite{chen2022visuo},  stage-guided fusion~\cite{feng2024play}, cross-modal attention mechanisms, and unified modality methods~\cite{yuan2023robot, huang20243dvitac}. 
However, all these approaches share a common limitation: they perform \textit{feature-level} fusion to create a single conditional input for a monolithic policy. This makes them vulnerable to sparsity, where learning becomes biased toward statistically dominant but contextually irrelevant modalities (e.g., vision) while discarding essential but sparse signals (e.g., touch). Our work addresses this issue by factorizing policies into modality-specific modules and composing complete action distributions at the \textit{policy level}.

\noindent\textbf{Compositional Generation and Energy-Based Models.} 
Our framework builds on compositional modeling, particularly the principle of combining simple models as a product of distributions to form a more expressive joint distribution~\cite{du2024compositional,liu2021learning,liu2022compositional,urain2021composable}. This idea is central to Energy-Based Models (EBMs), where summing energy functions corresponds to multiplying probability distributions~\cite{du2020compositional,gkanatsios2023energybased}. 
Diffusion models can be viewed as score-matching EBMs~\cite{song2021scorebased}, making their score functions naturally composable in the same way. 
Beyond image generation, compositionality has been applied to trajectory modeling~\cite{janner2022planning,ajay2022conditional_generative_modeling}, language models~\cite{du2023improving}, robotic planning with constraints~\cite{yang2023compositional,mishra2023generative}, and hierarchical planning~\cite{ajay2023compositional}. More broadly, compositional generative models extend to domains such as traffic generation~\cite{lin2024causal} and human motion~\cite{shafir2023human,sun2024coma}, highlighting the versatility of compositionality as a modeling principle. 
In robotics, related work has explored combining policies across heterogeneous sources such as simulation and real-world data~\cite{wang2024poco}. 
In contrast, our approach focuses on compositional learning across diverse modalities, enabling adaptive integration of modalities such as vision, touch, and semantic features within a unified policy. \looseness=-1

\noindent\textbf{Diffusion Models in Robot Learning.} 
Diffusion models have emerged as state-of-the-art policy representations in robotics, capable of modeling complex, multimodal action spaces. They have been successfully applied to imitation learning for single-arm~\cite{chi2023diffusionpolicy, reuss2023goal,zhu2025touch} and bimanual manipulation~\cite{chen2025bimanual}, tool use~\cite{chen2025toolasinterface}, motion planning~\cite{janner2022planning, saha2023edmp}, and reinforcement learning~\cite{ajay2022conditional_generative_modeling}. Extensions of diffusion planning incorporate hierarchical structures~\cite{li2023hierarchical}, and domain-specific adaptations such as autonomous driving~\cite{liao2024diffusiondrive}. Concurrent work by Zhang et al.~\cite{zhang2025compositional} explores composition of RGB and point cloud policies using fixed, manually-tuned weights, but is limited to visual modalities without adaptive weighting. Our framework introduces a structured approach to compositional learning that accommodates arbitrary modalities, including tactile feedback and semantic features, through learned context-dependent routing for adaptive modality weighting. \looseness=-1


\section{Approach}

\begin{figure}[tbp]
 \centering
 \includegraphics[width=0.5\textwidth]{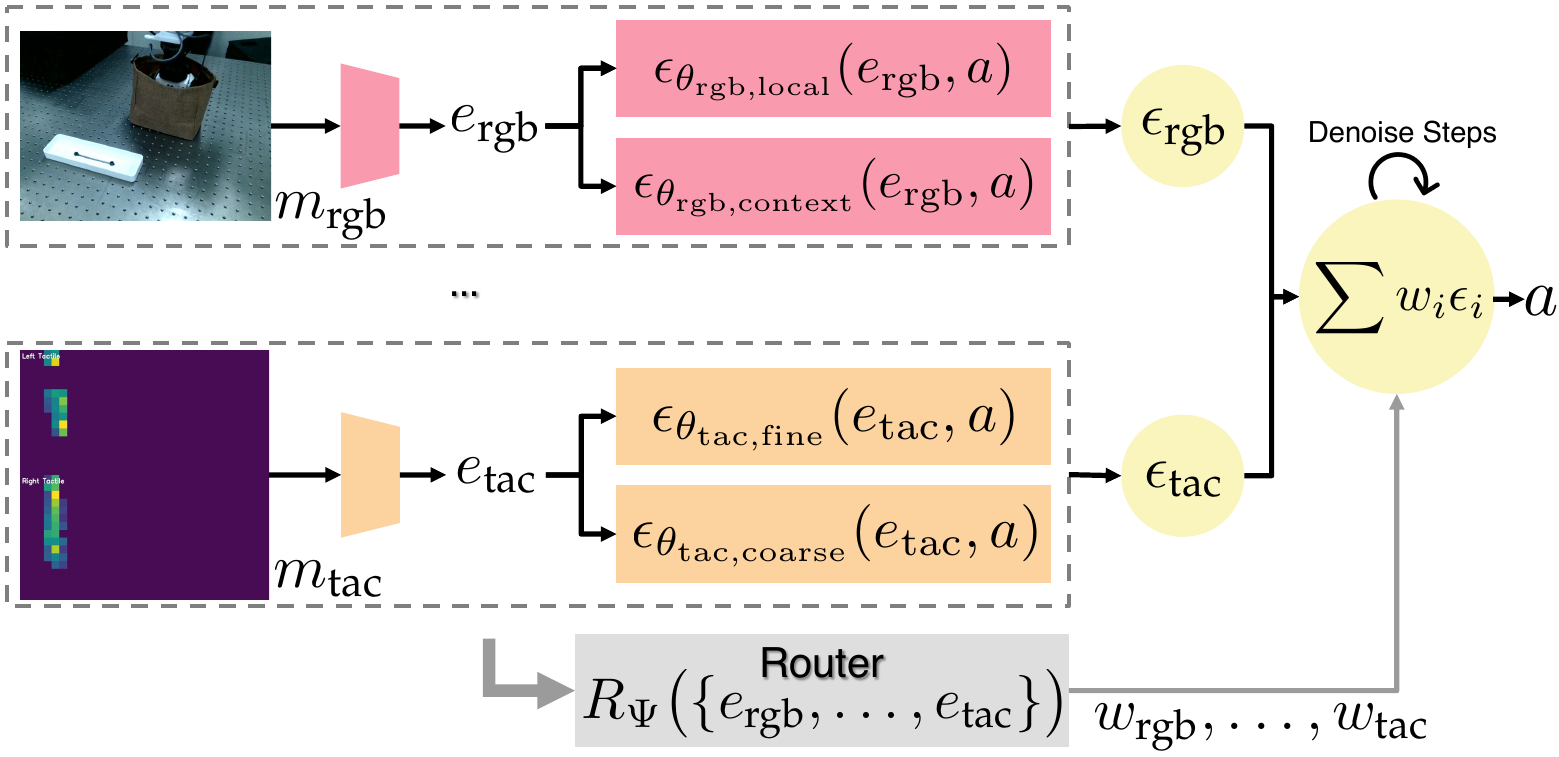}
 \vspace{-1pt}
\caption{\textbf{Overview of Our Compositional Policy Framework.} 
Raw sensory modalities ($m_{\text{rgb}}, m_{\text{tac}}$) are encoded into embeddings 
($\mathbf{e}_{\text{rgb}}, \mathbf{e}_{\text{tac}}$). 
Each modality is factorized into complementary sub-policies 
(e.g., $\epsilon_{\theta_{\text{rgb,context}}}(e_{\text{rgb}}, a)$, 
$\epsilon_{\theta_{\text{rgb,local}}}(e_{\text{rgb}}, a)$, 
$\epsilon_{\theta_{\text{tac,coarse}}}(e_{\text{tac}}, a)$, 
$\epsilon_{\theta_{\text{tac,fine}}}(e_{\text{tac}}, a)$), 
which produce score predictions that are averaged into a modality-specific score. 
A router network $R_\psi(\mathbf{e}_{\text{rgb}}, \dots, \mathbf{e}_{\text{tac}})$ 
then predicts consensus weights $\{w_i\}$ to reconcile these modality-specific scores 
into the final composed score $\sum_i w_i \epsilon_i$, which defines the policy for action generation.}
\vspace{-5pt}
 \label{fig:method}
\end{figure}

Our work introduces a compositional approach to multi-modal robot learning that addresses sparsity in sensory modalities through policy factorization and consensus-based composition (see Figure~\ref{fig:method}). We ground our framework in energy-based composition principles and instantiate them via policy consensus across modalities for robotic manipulation. \looseness=-1

\subsection{Problem Formulation}
We consider the imitation learning setting for robot manipulation. Given a dataset of expert demonstrations 
$\mathcal{D} = \{(\mathbf{s}_t, \mathbf{a}_t)\}_{t=1}^T$, 
where $\mathbf{s}_t$ is the state and $\mathbf{a}_t$ the action at timestep $t$, the state comprises $N$ sensory modalities:
\[
\mathbf{s}_t = \{m_{1,t}, m_{2,t}, \dots, m_{N,t}\}.
\]
Our goal is to learn a policy $\pi(\mathbf{a}_t | \mathbf{s}_t)$ that leverages heterogeneous information within $\mathbf{s}_t$. 
Each raw modality $m_{i,t}$ is encoded into a latent embedding $\mathbf{e}_{i,t}$ through a modality-specific encoder. 
Here, an embedding $\mathbf{e}_{i,t}$ refers to the modality encoding together with relevant robot state information (e.g., joint angles, gripper status), providing a richer context for action prediction.  
We denote $\mathbf{a}_t$ as the ground-truth action in the dataset, $a$ as a generic action candidate, and $a^k$ as its noised version at diffusion timestep $k$. 
For each modality $i$, we train $K_i$ sub-policies $p_{i,j}$, parameterized by $\theta_{i,j}$, which define energy functions $E_{\theta_{i,j}}(a, m_{i,t})$ and corresponding diffusion scores $\epsilon_{\theta_{i,j}}$. 
In our experiments, we set $K_i = 2$ to capture complementary behavioral modes 
(e.g., geometry vs. fine detail for vision, contact onset vs. sustained force for tactile), 
although the formulation allows general $K_i$.
A router network $R_\psi$ maps the embeddings $\{\mathbf{e}_{i,t}\}$ to consensus weights $\{w_i\}$, normalized by softmax so that $\sum_i w_i = 1$, indicating the relative influence of each modality. \looseness=-1


\subsection{Energy-Based Policy Composition}
Our approach builds on energy-based composition principles~\cite{liu2022compositional, du2020compositional, du2024reducereuserecyclecompositional, wang2024poco}, where policies are viewed as energy functions that assign low energy to preferred actions and high energy otherwise. Each base policy $p_{i,j}(a|m_{i,t})$ implicitly defines an energy function through
\[
p_{i,j}(a|m_{i,t}) \propto \exp(-E_{\theta_{i,j}}(a, m_{i,t})).
\]

Composing such policies corresponds to multiplying distributions, or equivalently summing their energies:
\[
p(a | \mathbf{s}_t) \propto \prod_{i=1}^{N} \left[\prod_{j=1}^{K_i} \exp(-E_{\theta_{i,j}}(a, m_{i,t}))\right]^{w_i},
\]
\[
= \exp\!\left(-\sum_{i=1}^{N} w_i \sum_{j=1}^{K_i} E_{\theta_{i,j}}(a, m_{i,t})\right).
\]

This reveals that our composition performs a weighted sum of energy functions, where the router weights $\{w_i\}$ determine each modality’s influence based on the current state. Unlike feature concatenation, which forces all modalities through a shared network that tends to suppress statistically rare signals, our formulation preserves separate energy functions for each modality, ensuring that sparse but critical signals remain influential.




\subsection{Compositional Policy Factorization}
We factorize the policy at two levels to capture both inter- and intra-modality structure:
\[
p(a|\mathbf{s}_t) \propto \prod_{i=1}^N p_i(a|m_{i,t})^{w_i},
\]
\[
p_i(a|m_{i,t}) \propto \prod_{j=1}^{K_i} p_{i,j}(a|m_{i,t}).
\]

Here $p_i$ is the composite policy for modality $i$, while $p_{i,j}$ are complementary sub-policies (e.g., vision experts for geometry vs. fine detail, tactile experts for contact onset vs. sustained force).  

This product-of-distributions view admits a constraint satisfaction interpretation~\cite{yang2023compositional}, where each modality-specific policy imposes behavioral constraints on the final action (e.g., geometry from vision, contact dynamics from touch). Importantly, unlike feature concatenation where sparse signals compete with dominant ones for network capacity, each $p_i$ is trained independently on its own modality stream. This ensures that rarely active but task critical modalities, such as tactile inputs that appear only during contact, retain their influence through the learned consensus weights $w_i$, which provide a consistent balancing of modality influence across the policy. \looseness=-1




\subsection{Score-Based Implementation via Diffusion Models}
We implement each base policy $p_{i,j}$ using Denoising Diffusion Probabilistic Models (DDPMs)~\cite{ho2020denoising}. Diffusion models can be interpreted as score-matching energy-based models~\cite{song2021scorebased}, and following compositional generation principles~\cite{du2020compositional, liu2022compositional}, sampling from a product of distributions corresponds to summing their score functions. This leads to a two-step aggregation process at each denoising step $k$:

\textit{Intra-Modality Composition.} The composed score for modality $i$ is obtained by averaging scores of its $K_i$ factorized sub-policies:
\[
\epsilon_{i,\text{comp}}(a^k, m_{i,t}, k) = \tfrac{1}{K_i}\sum_{j=1}^{K_i} \epsilon_{\theta_{i,j}}(a^k, m_{i,t}, k).
\]

\textit{Inter-Modality Composition.} The final composed score is then computed using router-weighted combination:
\[
\epsilon_{\text{comp}}(a^k, \mathbf{s}_t, k) = \sum_{i=1}^{N} w_i(\mathbf{s}_t) \cdot \epsilon_{i,\text{comp}}(a^k, m_{i,t}, k).
\]


The modality-specific scores $\epsilon_{i,\text{comp}}$ define gradient fields that encode each modality’s behavioral constraints. The router assigns consensus weights $w_i(\mathbf{s}_t)$ to combine these fields into a unified score, balancing their contributions. This weighted composition connects energy based composition with score based diffusion, establishing a direct theoretical link between compositional energy models and diffusion-based policies. \looseness=-1




\subsection{Router Network}
The router $R_\psi$ maps modality-specific embeddings $\{\mathbf{e}_i\}$ to consensus weights $\{w_i\}$, normalized with a softmax so that they are positive and sum to one, making them interpretable as relative modality influence. Conceptually, the router reconciles the action proposals of modality-specific experts into a unified policy. The consensus weights are learned during training and fixed at execution, providing a interpretable mechanism for balancing modalities.


\subsection{Advantages Over Existing Fusion}
Our framework offers  
(1) \textbf{Robustness to sparsity}, as rarely active modalities such as tactile remain represented by separate experts rather than being suppressed.
(2) \textbf{Modularity}, as experts can be trained and extended independently without retraining the entire policy.  
(3) \textbf{Interpretability}, with consensus weights $\{w_i\}$ directly revealing the influence of each modality.  
(4) \textbf{Principled consensus}, as the router provides a consistent weighting scheme grounded in energy-based composition, rather than blind feature fusion.

These properties make our framework a theoretically grounded and extensible alternative to monolithic feature concatenation for multi-modal robot learning.



\begin{figure*}[t!]
\centering
\includegraphics[width=\linewidth]{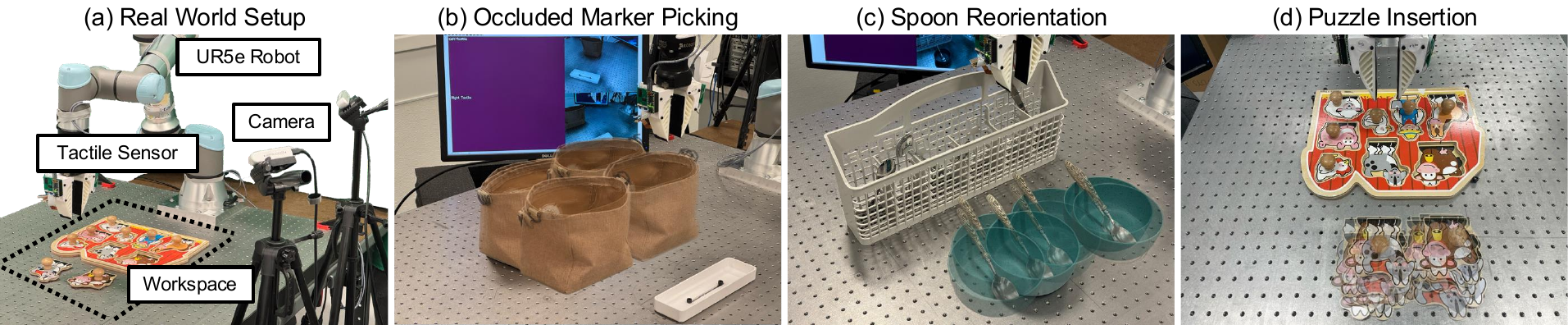}
\vspace{-5pt}
\caption{\textbf{Real-World Experimental Setup.} (a) UR5e manipulator equipped with dual cameras and tactile sensors. (b–d) Overlays of initial conditions for the evaluation tasks: occluded marker picking, spoon reorientation, and puzzle insertion.}
\label{fig:real_world_setup}
\vspace{-10pt}
\end{figure*}

\section{Experiment}

We evaluate our modality-composable policy framework by addressing three key research questions: (1) How does a compositional architecture compare to feature-level fusion in tasks with modality sparsity? (2) Does the compositional model capture context-dependent and modified reliance on different modalities, allowing new experts to be composed without retraining? (3) Does the policy maintain robustness under physical perturbations and sensor corruption? \looseness=-1

\begin{figure*}[t!]
\centering
\includegraphics[width=0.99\textwidth]{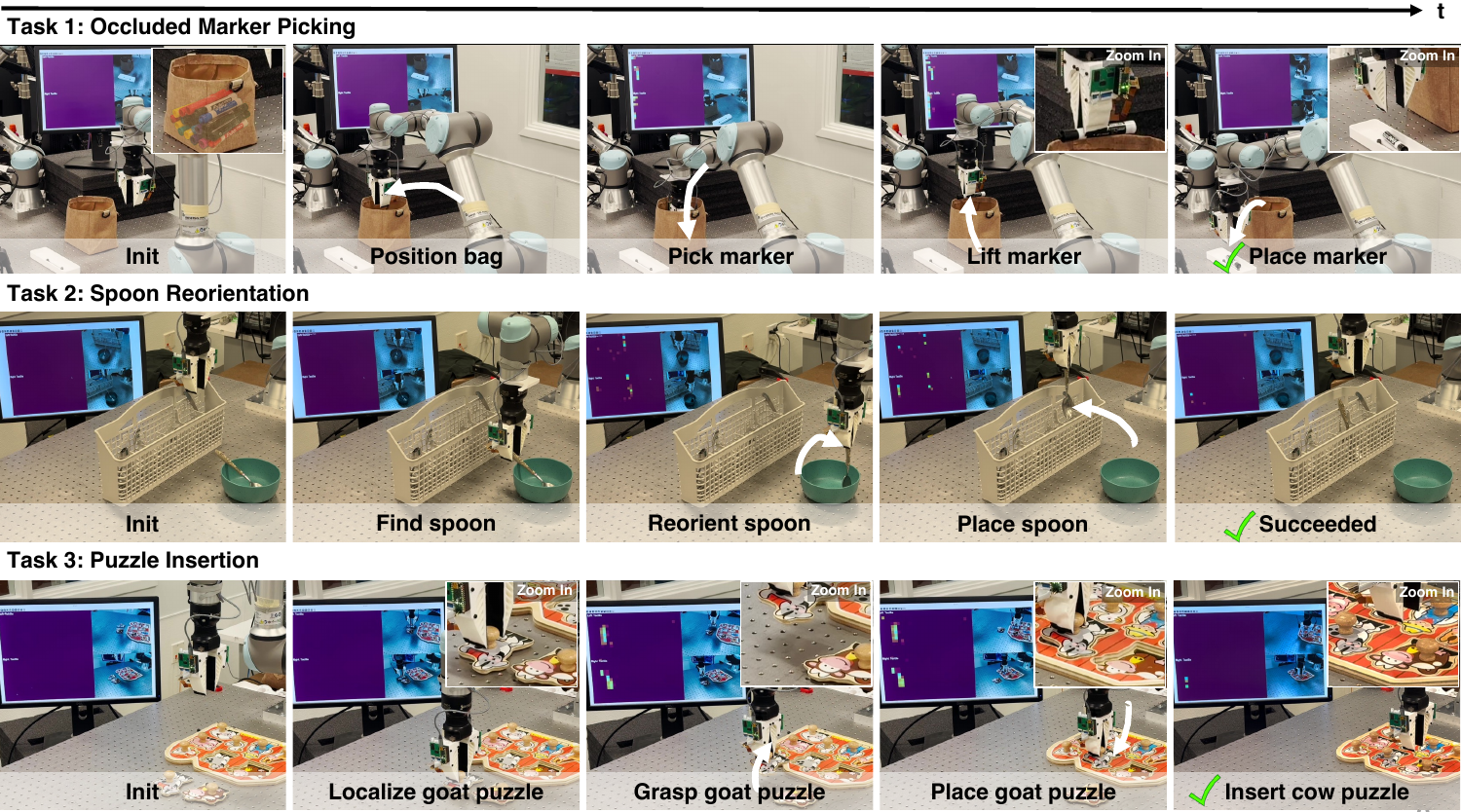}
\vspace{-1pt}
\caption{\textbf{Qualitative Policy Rollouts.} Representative execution traces from three tasks: 
Task 1 occluded marker picking, where tactile feedback guides manipulation when vision is unavailable; 
Task 2 spoon reorientation, demonstrating dexterous in-hand manipulation; 
Task 3 puzzle insertion, requiring high-precision alignment at millimeter accuracy.}
\label{fig:policy_rollout}
\vspace{-1pt}
\end{figure*}

\subsection{Experimental Setup}

\textbf{Tasks.} We evaluate our method on both simulation and real-world manipulation tasks.  
In simulation, we use RLBench~\cite{james2019rlbenchbenchmark} as a multi-task benchmark with four tasks: \textit{open box}, \textit{open drawer}, \textit{take umbrella out of stand}, and \textit{toilet seat up}. The policy is trained on a total of 200 demonstration episodes and evaluated on 200 unseen configurations.

For real-world experiments, we employ a UR5e manipulator equipped with dual cameras and tactile sensors, as shown in Figure~\ref{fig:real_world_setup}(a). We evaluate three challenging manipulation tasks, with Figure~\ref{fig:real_world_setup}(b–d) showing overlays of their initial testing conditions: (i) \textit{occluded marker picking}; (ii) \textit{spoon reorientation}; and (iii) \textit{puzzle insertion}. We collect 80, 60, and 50 teleoperated demonstrations for these tasks, respectively. \looseness=-1

\begin{figure*}[tbh]
\centering
\includegraphics[width=\linewidth]{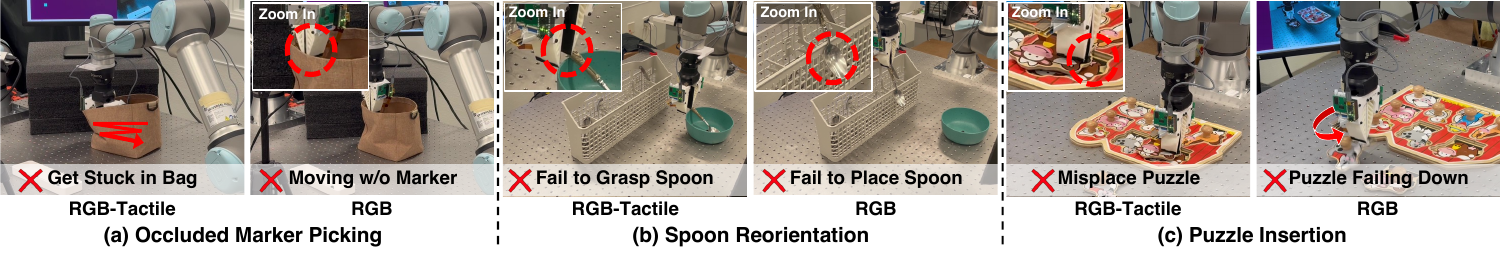}
\vspace{-10pt}
\caption{\textbf{Typical Failure Cases of Baseline Methods.}  
We show failure cases of an RGB-only policy compared with an RGB+Tactile concatenation baseline.  
Each task highlights the complementary roles of the two modalities: vision provides global spatial and geometric information, while tactile sensing provides contact awareness and fine-grained grasp feedback.  
(a) In occluded marker picking, the concatenation baseline becomes trapped without grasping, while RGB-only lacks awareness of the grasp state once occluded.  
(b) In spoon reorientation, the concatenation baseline fails at initial grasping, while RGB-only fails at precise placement.  
(c) In puzzle insertion, the concatenation baseline causes misalignment, while RGB-only suffers frequent grasp failures.}
\label{fig:baseline_failure}
\vspace{-5pt}
\end{figure*}

\textbf{Sensory Modalities.} 
In simulation, we utilize RGB images from two cameras, point cloud (PCD) data from depth sensors, and 3D semantic features extracted using a pretrained DINO model~\cite{caron2021emerging}. For real-world experiments, we use RGB images from dual side-view Intel RealSense D415 cameras with a resolution of $96 \times 128$, along with dense tactile arrays from \textit{FlexiTac} sensors~\cite{huang20243dvitac}. Each finger is equipped with a tactile pad consisting of $12 \times 32$ sensing units, each with a spatial resolution of $2\,\text{mm}$. \looseness=-1

\textbf{Baselines.} We compare against comprehensive baselines tailored to each domain:
\begin{itemize}
    \item \textbf{Simulation:} (i) Single-modality policies trained on RGB-only, PCD-only, or DINO-only inputs; (ii) Feature concatenation combining all modality embeddings; (iii) Factorized MoE fusion using soft routing.
    \item \textbf{Real-world:} (i) RGB-only policy; (ii) RGB+Tactile feature concatenation baseline.
\end{itemize}

\textbf{Metrics.} We report success rate as the primary metric and completion time as a secondary metric for successful trials.

\subsection{Main Results}

\textbf{Simulation Performance.} Table~\ref{table:sim_results} presents our simulation results. Our method achieves the highest average success rate (66\%), significantly outperforming both single-modality policies and the feature concatenation baseline (56\%). This 18\% relative improvement is achieved with minimal parameter overhead  (+0.7M parameters, where M denotes millions, corresponding to only a 0.3\% increase), demonstrating the efficiency of our compositional approach compared to naive fusion strategies.

\begin{table}[h!]
\centering
\caption{Policy performance and parameter count on RLBench tasks. Our method achieves 18\% relative improvement over concatenation baseline with negligible parameter increase.}
\label{table:sim_results}
\begin{tabular}{lcc}
\toprule
\textbf{Method} & \textbf{Success Rate} & \textbf{Params (M)} \\
\midrule
\textit{Single-Modality} & & \\
\quad RGB only & 0.54 & 257.3 \\
\quad Point Cloud only & 0.49 & 251.9 \\
\quad 3D DINO only & 0.45 & 251.9 \\
\midrule
\textit{Multi-Modality} & & \\
\quad Concatenation & 0.56 & 262.9 \\
\midrule
\textbf{Ours} & \textbf{0.66} & \textbf{263.6} \\
\bottomrule
\end{tabular}
\vspace{-10pt}
\end{table}

\textbf{Real-World Performance.} Tables~\ref{table:puzzle_results}--\ref{table:spoon_results} demonstrate consistent superiority across all real-world tasks. In contrast to baselines, our method achieves the highest success rates (65\% for occluded picking, 75\% for spoon reorientation, and 52\% for puzzle insertion) while maintaining the lowest average completion times. Figure~\ref{fig:policy_rollout} provides qualitative evidence of these performance differences.  

Notably, the RGB+Tactile concatenation baseline exhibits catastrophic failure: in \textit{Occluded Marker Picking}, its success rate is only 5\% compared to 35\% for RGB-only, and in \textit{Spoon Reorientation}, it achieves just 21\% compared to 67\% for RGB-only. These results confirm that naive fusion can be actively detrimental when modalities have varying informativeness. Figure~\ref{fig:baseline_failure} illustrates these systematic failure modes: the concatenation baseline gets trapped without successful grasping in occluded picking, fails at initial grasping in spoon reorientation, and causes misalignment in puzzle insertion, while RGB-only policies suffer from lack of tactile feedback during critical manipulation phases. \looseness=-1

\begin{table}[h!]
\centering
\caption{Decomposed Success Rates on Real-World Puzzle Insertion Task.}
\label{table:puzzle_results}
\begin{tabular}{l|ccc}
\toprule
\textbf{Sub-Task} & \textbf{RGB} & \textbf{RGB+Tac.} & \textbf{Ours} \\
\midrule
Pick up goat & 0.90 & 0.90 & \textbf{1.00} \\
Place goat & 0.90 & 0.90 & \textbf{0.95} \\
Task success (goat) & 0.25 & 0.35 & \textbf{0.58} \\
\midrule
Pick up cow & 0.85 & 0.80 & \textbf{0.95} \\
Place cow & 0.75 & 0.80 & \textbf{0.95} \\
Task success (cow) & 0.30 & \textbf{0.45} & \textbf{0.45} \\
\bottomrule
\end{tabular}
\vspace{-10pt}
\end{table}

\begin{table}[h!]
\centering
\caption{Performance on Occluded Marker Picking. Failed trials counted as 120s timeout.}
\label{table:occluded_results}
\begin{tabular}{l|ccc}
\toprule
\textbf{Metric} & \textbf{RGB} & \textbf{RGB+Tac.} & \textbf{Ours} \\
\midrule
Success Rate & 0.35 & 0.05 & \textbf{0.65} \\
Avg. Time (s) & 107.8 & 117.9 & \textbf{96.5} \\
\bottomrule
\end{tabular}
\vspace{-10pt}
\end{table}

\begin{table}[h!]
\centering
\caption{Performance on Spoon Reorientation. Failed trials counted as 300s timeout.}
\label{table:spoon_results}
\begin{tabular}{l|ccc}
\toprule
\textbf{Metric} & \textbf{RGB} & \textbf{RGB+Tac.} & \textbf{Ours} \\
\midrule
Success Rate & 0.67 & 0.21 & \textbf{0.75} \\
Avg. Time (s) & 117.1 & 221.5 & \textbf{94.8} \\
\bottomrule
\end{tabular}
\vspace{-10pt}
\end{table}

\begin{figure*}[tbp]
\centering
\includegraphics[width=\textwidth]{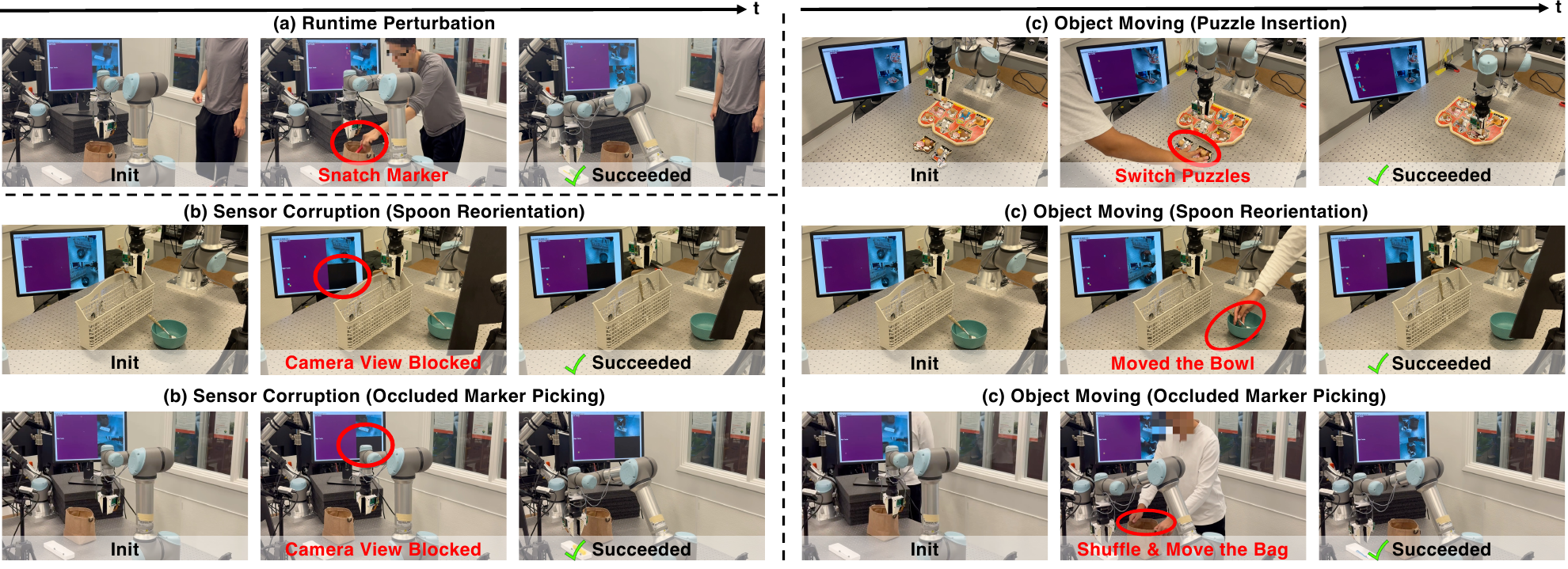}
\vspace{-10pt}
\caption{\textbf{Policy Robustness under Diverse Perturbations.} 
We evaluate three types of interventions: 
(a) runtime perturbation, where the marker is suddenly snatched away during execution; 
(b) sensor corruption, where a camera is occluded to simulate partial sensor failure; and 
(c) object repositioning, where task-relevant objects are reset and moved to new positions between executions. 
Our method maintains reliable performance across all scenarios.}
\label{fig:robustness_analysis}
\vspace{-5pt}
\end{figure*}


\subsection{Analysis of Learned Modality Dependencies}
To assess whether the policy leverages different modalities depending on task context, we conduct a perturbation-based importance analysis. Specifically, we measure modality dependency by injecting calibrated Gaussian noise $\mathcal{N}(0, \sigma^2)$ into each modality and computing the normalized L2 distance between perturbed and original action outputs. For stability, temporal smoothing is applied using an exponential moving average (EMA, $\alpha=0.1$).

As shown in Figure~\ref{fig:teaser}a, this analysis reveals context aware modality usage during the \textit{Occluded Marker Picking} task, which unfolds in two distinct phases. In the first stage (t = 0 to 5 steps), vision dominates: the policy shows high sensitivity to visual perturbations, relying on visual feedback for spatial navigation and approach. In the second stage (after 5 steps), the policy shifts to multi modal coordination. As the gripper enters the occluded container, tactile sensitivity rises sharply upon contact while vision remains important, highlighting their complementary roles. This emergent two stage behavior demonstrates that our compositional architecture can transition from single modality reliance to multi modal integration based on task demands. \looseness=-1

\begin{figure*}[tbp]
\centering
\includegraphics[width=\textwidth]{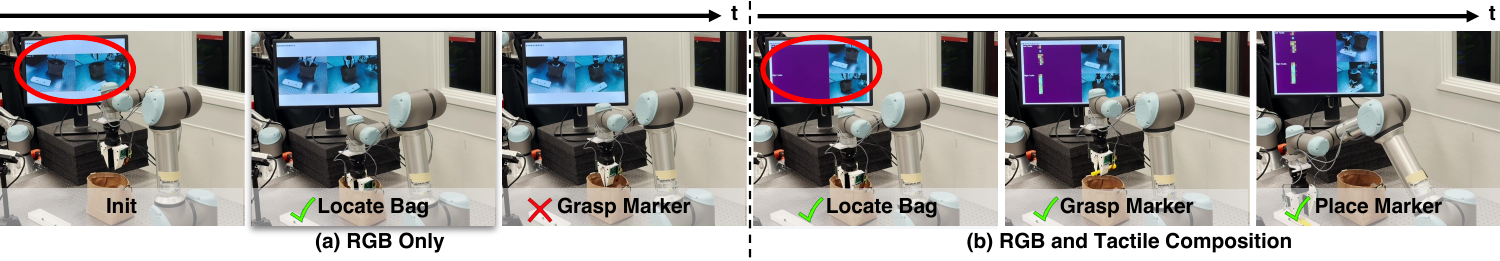}
\vspace{-10pt}
\caption{\textbf{Incremental Learning.} 
(a)~An RGB-only policy fails to grasp the marker without tactile feedback. 
(b)~By composing a pre-trained RGB policy with a tactile policy using manually set consensus weights $(0.5, 0.5)$, the combined policy successfully grasps the marker under occlusion without requiring retraining.}
\label{fig:incremental_learning}
\vspace{-5pt}
\end{figure*}

\subsection{Robustness and Adaptation Evaluation}

We evaluate the robustness and adaptation of our approach through systematic perturbation experiments across physical and sensory domains. \looseness=-1

\textbf{Physical Perturbations.} We introduce two categories of physical interventions to assess policy resilience under environmental disturbances (Figure~\ref{fig:robustness_analysis}a, c):  
(a)~\emph{Runtime perturbation}: a mid-execution intervention in which the marker is suddenly snatched away during the occluded marker picking task, introducing a dynamic disturbance.
(c)~\emph{Object repositioning}: objects are deliberately reset and moved to new positions between task executions (e.g., relocating the bag in the occluded-marker task, displacing the bowl and spoon in the spoon-reorientation task, or swapping the positions of the goat and cow puzzle pieces).

\textbf{Sensor Corruption.} We simulate partial sensor failure by occluding one camera with an opaque card, eliminating visual input (Figure~\ref{fig:robustness_analysis}b). This corruption is applied to both spoon-reorientation and occluded-marker tasks. Our method maintains consistent success despite the degraded sensory input, demonstrating resilience to partial sensor failure.

\textbf{Incremental Learning.} 
A key benefit of our compositional framework is the ability to combine independently trained policies without retraining. 
We demonstrate this by composing an RGB policy and a tactile policy using fixed consensus weights $(0.5, 0.5)$. 
As shown in Figure~\ref{fig:incremental_learning}, while the RGB-only policy fails under occlusion, the composed policy succeeds in grasping despite never being jointly trained. 
This highlights that our method supports incremental integration of new sensory modalities, enabling efficient extension of existing skills. \looseness=-1


\subsection{Ablation Studies}
\label{subsec:ablation}

To validate our key architectural choices, we conduct comprehensive ablation studies examining the impact of different routing and fusion strategies. Table~\ref{table:ablation_results} summarizes the experimental results.

\textbf{Learned vs. Fixed Consensus.} 
We first investigate the contribution of our learned consensus mechanism by replacing it with a fixed equal-weight strategy. This ablation results in a significant performance degradation of 7.6\% (from 66\% to 61\%), demonstrating that adaptive, learned consensus weighting is important for effective policy composition. \looseness=-1

\textbf{Policy-Level vs. Feature-Level Fusion.} 
We compare our policy-level consensus-based composition against a soft-routing MoE baseline that performs fusion at the feature level. Our method achieves a substantial improvement of 15.8\% over the MoE baseline (66\% vs. 57\%; see Table~\ref{table:ablation_results}). This performance gap suggests that composing modalities at the action-output level better preserves modality-specific information compared to early feature selection, which may prematurely discard relevant signals needed for downstream decision-making. \looseness=-1

\begin{table}[ht]
\centering
\caption{Ablation study comparing routing and fusion strategies. Results show average success rates across all evaluation tasks in simulation.}
\label{table:ablation_results}
\begin{tabular}{lc}
\toprule
\textbf{Method} & \textbf{Avg. Success Rate} \\
\midrule
\textbf{Ours (Learned Router)} & \textbf{0.66} \\
\midrule
\multicolumn{2}{l}{\textit{Ablation Variants}} \\
\quad Fixed Equal Weights & 0.61  \\
\quad Factorized MoE Fusion & 0.57 \\
\bottomrule
\end{tabular}
\vspace{-0.5em}
\end{table}

\section{Conclusion}

We presented a compositional framework for multimodal robot manipulation that factorizes policies at the modality level. The action distribution conditioned on each input modality is modeled by a separate expert policy, and their outputs are integrated through learned consensus weights from a router network. This consensus-based composition allows the policy to adaptively balance sensing modalities, preserving their expertise while enabling robust coordination. 
Experiments on a multi-task simulation benchmark and contact-rich real-world tasks demonstrate that our method consistently outperforms conventional feature-fusion baselines. Beyond these performance gains, our importance analysis reveals that policies dynamically shift reliance between modalities based on context, with vision handling geometric reasoning and tactile managing contact-rich phases. These compositional advantages enable incremental sensor deployment without retraining and provide natural robustness to sensor failures.

While promising, our framework opens several directions for future research. First, the current router adapts consensus weights at the policy level but does not provide fine-grained or temporally dynamic adjustments; developing more expressive consensus mechanisms could enhance adaptability. Second, our experiments focus primarily on vision and tactile inputs in controlled real-world environments, leaving extensions to additional modalities (e.g., audio, force, language) and deployment in more diverse settings as important next steps. \looseness=-1




\section*{Acknowledgments}
{
We thank Ananya Yammanuru for assistance with the real-world robot setup, Jiaheng Han for insightful discussions on experiment design visualization, Shaoxiong Yao and Wei-Cheng Huang for their contributions to method presentation, and Ye-Ji Mun and Shaoting Peng for valuable feedback on paper drafts. This material is based upon work supported by the National Science Foundation under Grant No. 2143435. We also thank PickleRobotics for their generous support of the lab. \looseness=-1
}

\bibliographystyle{IEEEtran}
\bibliography{BibFile}

@inproceedings{caron2021emerging,
  title={Emerging Properties in Self-Supervised Vision Transformers},
  author={Caron, Mathilde and Touvron, Hugo and Misra, Ishan and J\'egou, Herv\'e  and Mairal, Julien and Bojanowski, Piotr and Joulin, Armand},
  booktitle={Proceedings of the International Conference on Computer Vision (ICCV)},
  year={2021}
}

@inproceedings{chang2020sound,
author = {Chang, Peixin and Liu, Shuijing and Chen, Haonan and Driggs-Campbell, Katherine},
title = {Robot Sound Interpretation: Combining Sight and Sound in Learning-Based Control},
year = {2020},
publisher = {IEEE Press},
booktitle = {2020 IEEE/RSJ International Conference on Intelligent Robots and Systems (IROS)},
pages = {5580–5587},
numpages = {8},
location = {Las Vegas, NV, USA}
}

@inproceedings{
feng2024play,
title={Play to the Score: Stage-Guided Dynamic Multi-Sensory Fusion for Robotic Manipulation},
author={Ruoxuan Feng and Di Hu and Wenke Ma and Xuelong Li},
booktitle={8th Annual Conference on Robot Learning},
year={2024},
}

@article{Huang2023TowardsSM,
  title={Towards Safe Multi-Level Human-Robot Interaction in Industrial Tasks},
  author={Zhe Huang and Ye-Ji Mun and Haonan Chen and Yiqing Xie and Yilong Niu and Xiang Li and Ninghan Zhong and Ha-Il You and David Livingston McPherson and K. Driggs-Campbell},
  journal={ArXiv},
  year={2023},
  volume={abs/2308.03222},
}

@article{liu2022compositional,
  title={Compositional Visual Generation with Composable Diffusion Models},
  author={Liu, Nan and Li, Shuang and Du, Yilun and Torralba, Antonio and Tenenbaum, Joshua B},
  journal={arXiv preprint arXiv:2206.01714},
  year={2022}
}

@article{urain2021composable,
author = {Julen Urain and Anqi Li and Puze Liu and Carlo D’Eramo and Jan Peters},
title ={Composable energy policies for reactive motion generation and reinforcement learning},

journal = {The International Journal of Robotics Research},
volume = {42},
number = {10},
pages = {827-858},
year = {2023},
doi = {10.1177/02783649231179499},


}

@inproceedings{huang20243dvitac,
    title={3D ViTac:Learning Fine-Grained Manipulation with Visuo-Tactile Sensing},
    author={Huang, Binghao and Wang, Yixuan and Yang, Xinyi and Luo, Yiyue and Li, Yunzhu},
    booktitle={Proceedings of Robotics: Conference on Robot Learning(CoRL)},
        year={2024}
}

@misc{james2019rlbenchbenchmark,
      title={RLBench: The Robot Learning Benchmark \& Learning Environment}, 
      author={Stephen James and Zicong Ma and David Rovick Arrojo and Andrew J. Davison},
      year={2019},
      eprint={1909.12271},
      archivePrefix={arXiv},
      primaryClass={cs.RO},
}

@misc{du2024reducereuserecyclecompositional,
      title={Reduce, Reuse, Recycle: Compositional Generation with Energy-Based Diffusion Models and MCMC}, 
      author={Yilun Du and Conor Durkan and Robin Strudel and Joshua B. Tenenbaum and Sander Dieleman and Rob Fergus and Jascha Sohl-Dickstein and Arnaud Doucet and Will Grathwohl},
      year={2024},
      eprint={2302.11552},
      archivePrefix={arXiv},
      primaryClass={cs.LG},
}

@article{wang2024poco,
  title={Poco: Policy composition from and for heterogeneous robot learning},
  author={Wang, Lirui and Zhao, Jialiang and Du, Yilun and Adelson, Edward H and Tedrake, Russ},
  journal={arXiv preprint arXiv:2402.02511},
  year={2024}
}

@inproceedings{reuss2023goal,
  title={Goal-Conditioned Imitation Learning using Score-based Diffusion Policies},
  author={Reuss, Moritz and Li, Maximilian and Jia, Xiaogang and Lioutikov, Rudolf},
  booktitle={Proceedings of Robotics: Science and Systems (RSS)},
  year={2023}
}

@article{du2024compositional,
  title={Compositional generative modeling: A single model is not all you need},
  author={Du, Yilun and Kaelbling, Leslie},
  journal={arXiv preprint arXiv:2402.01103},
  year={2024}
}

@inproceedings{ho2020denoising,
    author = {Ho, Jonathan and Jain, Ajay and Abbeel, Pieter},
    booktitle = {Advances in Neural Information Processing Systems},
    title = {Denoising Diffusion Probabilistic Models},
    year = {2020}
}

@inproceedings{du2020compositional,
  title={Compositional Visual Generation with Energy Based Models},
  author={Du, Yilun and Li, Shuang and Mordatch, Igor},
  booktitle={Advances in Neural Information Processing Systems},
  year={2020},
}

@inproceedings{lee2019multimodal,
author = {Lee, Michelle A. and Zhu, Yuke and Srinivasan, Krishnan and Shah, Parth and Savarese, Silvio and Fei-Fei, Li and Garg, Animesh and Bohg, Jeannette},
title = {Making Sense of Vision and Touch: Self-Supervised Learning of Multimodal Representations for Contact-Rich Tasks},
year = {2019},
publisher = {IEEE Press},
doi = {10.1109/ICRA.2019.8793485},
booktitle = {2019 International Conference on Robotics and Automation (ICRA)},
pages = {8943–8950},
numpages = {8},
location = {Montreal, QC, Canada}
}

@inproceedings{
  song2021scorebased,
  title={Score-Based Generative Modeling through Stochastic Differential Equations},
  author={Yang Song and Jascha Sohl-Dickstein and Diederik P Kingma and Abhishek Kumar and Stefano Ermon and Ben Poole},
  booktitle={International Conference on Learning Representations},
  year={2021},
}

@inproceedings{chi2023diffusionpolicy,
	title={Diffusion Policy: Visuomotor Policy Learning via Action Diffusion},
	author={Chi, Cheng and Feng, Siyuan and Du, Yilun and Xu, Zhenjia and Cousineau, Eric and Burchfiel, Benjamin and Song, Shuran},
	booktitle={Proceedings of Robotics: Science and Systems (RSS)},
	year={2023}
}

@inproceedings{janner2022planning,
  title = {Planning with Diffusion for Flexible Behavior Synthesis},
  author = {Michael Janner and Yilun Du and Joshua B. Tenenbaum and Sergey Levine},
  booktitle = {International Conference on Machine Learning},
  year = {2022},
}

@article{dave2024multimodal,
  title={Multimodal Visual-Tactile Representation Learning through Self-Supervised Contrastive Pre-Training},
  author={Dave, Vedant and Lygerakis, Fotios and Rueckert, Elmar},
  journal={arXiv preprint arXiv:2401.12024},
  year={2024}
}

@article{suresh2023neural,
title={{N}eural feels with neural fields: {V}isuo-tactile perception for in-hand manipulation},
author={Suresh, Sudharshan and Qi, Haozhi and Wu, Tingfan and Fan, Taosha and Pineda, Luis and Lambeta, Mike and Malik, Jitendra and Kalakrishnan, Mrinal and Calandra, Roberto and Kaess, Michael and Ortiz, Joseph and Mukadam, Mustafa},
journal={Science Robotics},
pages={adl0628},
year={2024},
publisher={American Association for the Advancement of Science}}

@article{Johansson2004RolesOG,
  title={Roles of glabrous skin receptors and sensorimotor memory in automatic control of precision grip when lifting rougher or more slippery objects},
  author={Roland S. Johansson and G{\"o}ran Westling},
  journal={Experimental Brain Research},
  year={2004},
  volume={56},
  pages={550-564},
}

@article{Bicchi2000RoboticGA,
  title={Robotic grasping and contact: a review},
  author={Antonio Bicchi and Vijay R. Kumar},
  journal={Proceedings 2000 ICRA. Millennium Conference. IEEE International Conference on Robotics and Automation. Symposia Proceedings (Cat. No.00CH37065)},
  year={2000},
  volume={1},
  pages={348-353 vol.1},
}

@inproceedings{chen2022visuo,
  title={Visuo-tactile transformers for manipulation},
  author={Chen, Yizhou and Van der Merwe, Mark and Sipos, Andrea and Fazeli, Nima},
  booktitle={6th Annual Conference on Robot Learning},
  year={2022}
}

@article{yuan2023robot,
  title={Robot synesthesia: In-hand manipulation with visuotactile sensing},
  author={Yuan, Ying and Che, Haichuan and Qin, Yuzhe and Huang, Binghao and Yin, Zhao-Heng and Lee, Kang-Won and Wu, Yi and Lim, Soo-Chul and Wang, Xiaolong},
  journal={arXiv preprint arXiv:2312.01853},
  year={2023}
}

@INPROCEEDINGS{johanna2022multimodal,
  author={Hansen, Johanna and Hogan, Francois and Rivkin, Dmitriy and Meger, David and Jenkin, Michael and Dudek, Gregory},
  booktitle={2022 International Conference on Robotics and Automation (ICRA)}, 
  title={Visuotactile-RL: Learning Multimodal Manipulation Policies with Deep Reinforcement Learning}, 
  year={2022},
  volume={},
  number={},
  pages={8298-8304},
  doi={10.1109/ICRA46639.2022.9812019}}

@inproceedings{gkanatsios2023energybased,
        title={{Energy-based Models are Zero-Shot Planners for Compositional Scene Rearrangement}},
        author={Gkanatsios, Nikolaos and Jain, Ayush and Xian, Zhou and Zhang, Yunchu and Atkeson, Christopher and Fragkiadaki, Katerina},
        booktitle={Robotics: Science and Systems},
        year={2023}
      }

@misc{zhang2025compositional,
      title={Modality-Composable Diffusion Policy via Inference-Time Distribution-level Composition}, 
      author={Jiahang Cao and Qiang Zhang and Hanzhong Guo and Jiaxu Wang and Hao Cheng and Renjing Xu},
      year={2025},
      eprint={2503.12466},
      archivePrefix={arXiv},
      primaryClass={cs.RO},
}

@article{saha2023edmp,
  title={EDMP: Ensemble-of-costs-guided Diffusion for Motion Planning},
  author={Saha, Kallol and Mandadi, Vishal and Reddy, Jayaram and Srikanth, Ajit and Agarwal, Aditya and Sen, Bipasha and Singh, Arun and Krishna, Madhava},
  journal={arXiv},
  year={2023}
}

@inproceedings{
liu2021learning,
title={Learning to Compose Visual Relations},
author={Nan Liu and Shuang Li and Yilun Du and Joshua B. Tenenbaum and Antonio Torralba},
booktitle={Advances in Neural Information Processing Systems},
editor={A. Beygelzimer and Y. Dauphin and P. Liang and J. Wortman Vaughan},
year={2021},
}

@article{ajay2022conditional_generative_modeling,
  title={Is conditional generative modeling all you need for decision-making?},
  author={Ajay, Anurag and Du, Yilun and Gupta, Abhi and Tenenbaum, Joshua and Jaakkola, Tommi and Agrawal, Pulkit},
  journal={arXiv preprint arXiv:2211.15657},
  year={2022}
}

@inproceedings{mishra2023generative,
  title={Generative skill chaining: Long-horizon skill planning with diffusion models},
  author={Mishra, Udit Arora and Xue, Shuchen and Chen, Yifan and Xu, Danfei},
  booktitle={Conference on Robot Learning},
  pages={2905--2925},
  year={2023},
  organization={PMLR}
}

@article{ajay2023compositional,
  title={Compositional foundation models for hierarchical planning},
  author={Ajay, Anurag and Han, Song and Du, Yilun and Li, Shuang and Gupta, Abhinav and Jaakkola, Tommi and Tenenbaum, Joshua and Kaelbling, Leslie and Srivastava, Akash and Agrawal, Pulkit},
  journal={arXiv preprint arXiv:2309.08587},
  year={2023}
}

@article{liao2024diffusiondrive,
  title={DiffusionDrive: Truncated Diffusion Model for End-to-End Autonomous Driving},
  author={Liao, Bencheng and Chen, Shaoyu and Yin, Haoran and Jiang, Bo and Wang, Cheng and Yan, Sixu and Zhang, Xinbang and Li, Xiangyu and Zhang, Ying and Zhang, Qian and others},
  journal={arXiv preprint arXiv:2411.15139},
  year={2024}
}

@inproceedings{li2023hierarchical,
  title={Hierarchical diffusion for offline decision making},
  author={Li, Wenhao and Wang, Xiangfeng and Jin, Bo and Zha, Hongyuan},
  booktitle={International Conference on Machine Learning},
  pages={20035--20064},
  year={2023}
}

@article{sun2024coma,
  title={CoMA: Compositional Human Motion Generation with Multi-modal Agents},
  author={Sun, Shanlin and De Araujo, Gabriel and Xu, Jiaqi and Zhou, Shenghan and Zhang, Hanwen and Huang, Ziheng and You, Chenyu and Xie, Xiaohui},
  journal={arXiv preprint arXiv:2412.07320},
  year={2024}
}

@article{shafir2023human,
  title={Human motion diffusion as a generative prior},
  author={Shafir, Yonatan and Tevet, Guy and Kapon, Roy and Bermano, Amit H},
  journal={arXiv preprint arXiv:2303.01418},
  year={2023}
}

@article{lin2024causal,
  title={Causal Composition Diffusion Model for Closed-loop Traffic Generation},
  author={Lin, Haohong and Huang, Xin and Phan-Minh, Tung and Hayden, David S and Zhang, Huan and Zhao, Ding and Srinivasa, Siddhartha and Wolff, Eric M and Chen, Hongge},
  journal={arXiv preprint arXiv:2412.17920},
  year={2024}
}

@article{du2023improving,
  title={Improving factuality and reasoning in language models through multiagent debate},
  author={Du, Yilun and Li, Shuang and Torralba, Antonio and Tenenbaum, Joshua B and Mordatch, Igor},
  journal={arXiv preprint arXiv:2305.14325},
  year={2023}
}

@inproceedings{chen2025bimanual,
        title={Learning Coordinated Bimanual Manipulation Policies using State Diffusion and Inverse Dynamics Models},
        author={Haonan Chen and Jiaming Xu and Lily Sheng and Tianchen Ji and Shuijing Liu and Yunzhu Li and Katherine Driggs-Campbell},
        booktitle={2025 IEEE International Conference on Robotics and Automation (ICRA)},  
        year={2025}
}

@inproceedings{chen2025toolasinterface,
      title={Tool-as-Interface: Learning Robot Policies from Observing Human Tool Use}, 
      author={Haonan Chen and Cheng Zhu and Shuijing Liu and Yunzhu Li and Katherine Driggs-Campbell},
      booktitle={Conference on Robot Learning (CoRL)},
      year={2025}
}

@InProceedings{li2023see,
  title = 	 {See, Hear, and Feel: Smart Sensory Fusion for Robotic Manipulation},
  author =       {Li, Hao and Zhang, Yizhi and Zhu, Junzhe and Wang, Shaoxiong and Lee, Michelle A and Xu, Huazhe and Adelson, Edward and Fei-Fei, Li and Gao, Ruohan and Wu, Jiajun},
  booktitle = 	 {Proceedings of The 6th Conference on Robot Learning},
  pages = 	 {1368--1378},
  year = 	 {2023},
  editor = 	 {Liu, Karen and Kulic, Dana and Ichnowski, Jeff},
  volume = 	 {205},
  series = 	 {Proceedings of Machine Learning Research},
  month = 	 {14--18 Dec},
  publisher =    {PMLR}
}

@InProceedings{yang2023compositional,
  title = 	 {Compositional Diffusion-Based Continuous Constraint Solvers},
  author =       {Yang, Zhutian and Mao, Jiayuan and Du, Yilun and Wu, Jiajun and Tenenbaum, Joshua B. and Lozano-P\'{e}rez, Tom\'{a}s and Kaelbling, Leslie Pack},
  booktitle = 	 {Proceedings of The 7th Conference on Robot Learning},
  pages = 	 {3242--3265},
  year = 	 {2023},
  editor = 	 {Tan, Jie and Toussaint, Marc and Darvish, Kourosh},
  volume = 	 {229},
  series = 	 {Proceedings of Machine Learning Research},
  month = 	 {06--09 Nov},
  publisher =    {PMLR}
}

@article{zhu2025touch,
  title={Touch in the Wild: Learning Fine-Grained Manipulation with a Portable Visuo-Tactile Gripper},
  author={Zhu, Xinyue and Huang, Binghao and Li, Yunzhu},
  journal={arXiv preprint arXiv:2507.15062},
  year={2025}
}
\clearpage

\end{document}